\documentclass{article}

\usepackage[preprint]{neurips_2024}

\usepackage[utf8]{inputenc} % allow utf-8 input
\usepackage[T1]{fontenc}    % use 8-bit T1 fonts
\usepackage{hyperref}       % hyperlinks
\usepackage{url}            % simple URL typesetting
\usepackage{booktabs}       % professional-quality tables
\usepackage{amsfonts}       % blackboard math symbols
\usepackage{nicefrac}       % compact symbols for 1/2, etc.
\usepackage{microtype}      % microtypography
\usepackage{xcolor}         % colors

\usepackage{graphicx}
\usepackage{amsmath}
\usepackage{amssymb}
\usepackage{bbm}
\usepackage{algorithm}
\usepackage{algorithmic}
\usepackage{multirow}
\usepackage{url}
\usepackage{float}
\usepackage{enumerate}
\usepackage{multicol}

\usepackage{makecell}

\title{Active-Passive Federated Learning for Vertically Partitioned Multi-view Data}

\author{%
  Jiyuan Liu$^1$, Siqi Wang$^1$, Xinhang Wan$^1$, Yi Zhang$^1$, Junsong Chen$^1$, Xin Lu$^1$ and Xinwang Liu$^1$\\
  $^1$National University of Defense Technology, Changsha, Hunan, China 410073 \\
  \texttt{liujiyuan13@nudt.edu.cn} \\
}

\begin{document}

\maketitle

\begin{abstract}
    Vertical federated learning is a natural and elegant approach to integrate multi-view data vertically partitioned across devices (clients) while preserving their privacies.
    Apart from the model training, existing methods requires the collaboration of all clients in the model inference.
    However, the model inference is probably maintained for service in a long time, while the collaboration, especially when the clients belong to different organizations, is unpredictable in real-world scenarios, such as concellation of contract, network unavailablity, etc., resulting in the failure of them. 
    To address this issue, we, at the first attempt, propose a flexible Active-Passive Federated learning (APFed) framework.
    Specifically, the active client is the initiator of a learning task and responsible to build the complete model, while the passive clients only serve as assistants.
    Once the model built, the active client can make inference independently.
    In addition, we instance the APFed framework into two classification methods with employing the reconstruction loss and the contrastive loss on passive clients, respectively.
    Meanwhile, the two methods are tested in a set of experiments and achieves desired results, validating their effectiveness.
\end{abstract}

\section{Introduction}
\label{sec:intro}

% FL intro
Federated Learning, firstly proposed by Google researchers in \cite{DBLP:conf/aistats/McMahanMRHA17}, is an emerging and promising approach to integrate information distributed across different clients.
It builds the machine learning model by exchanging the parameter gradients or data representations rather than data observations, thus can preserve the data privacies \cite{DBLP:journals/tist/YangLCT19,9866779,DBLP:conf/iclr/YuLWXL23,DBLP:conf/aaai/ThapaCCS22}.
As one of its most important branches, vertical federated learning is an elegant and effective approach to integarte the data of different features across devices (clients).
Currently, it has been widely applied in a large volume of academic and industrial scenarios, such as medical treatment \cite{DBLP:journals/tist/CheKPSLCH22}, recommendation system \cite{DBLP:conf/nips/CuiCLYW21}, finance \cite{DBLP:conf/kdd/0001ZWWFTWLWH21}, etc. 

Multi-view data refer to the multiple descriptions of a same set of data samples \cite{9845473,DBLP:journals/pami/LiuLYLX23,DBLP:journals/tkde/LiuLXLZWY22,DBLP:conf/icml/0001HLZZ19,DBLP:journals/tip/WangLLNL22}.
Usually, the descriptions can be recognized semantically.
For example, when banks generate the customer persona to make loans, they often collect its information of multiple aspects, i.e. multiple descriptions, such as historical credits, consumption data and medical records.
In past decades, a large number of researches were proposed by integrating the multi-view data to improve the performance of machine learning tasks, including classification \cite{DBLP:journals/pami/HanZFZ23,DBLP:journals/pami/ZhangCHZFH22,10102671}, clustering \cite{DBLP:journals/pami/ZhangLSSS19,DBLP:journals/tkde/HuangLTXL23,DBLP:journals/tcyb/RenYSW21,DBLP:journals/tkde/XuRTYPYPYH23,DBLP:journals/tip/WangLWZZH15,DBLP:journals/tip/LiTZLZZ22,DBLP:journals/tnn/LiuLYGKH22}, anomaly detection \cite{9810850}, etc.
Most of them assume that the multi-view data are centralized and fully accessible in model building.
However, the assumption does not always hold, especially when the data are collected from multiple sources of different organizations.
Taking the aforementioned customer persona as an example, the historical credits are possibly owned by banks, while the consumption are mostly collected by shopping platforms, and the medical records belong to hospitals.
It is unrealistic for these organizations to share the original data due to privacy and security concerns, leading to the impossibility of centralizing them.
By taking each organization as a client, vertical federated learning is an ideal approach to integrated such \textbf{vertically partitioned multi-view data}, and receives great attentions from researchers.
It is worth to note that, in literature, some researches label the vertical federated learning on multi-view data as federated multi-view learning \cite{DBLP:journals/tist/CheKPSLCH22,DBLP:journals/pr/HuangSXTL22}.

\begin{figure*}[t]
    \centering
    \includegraphics[width=\linewidth]{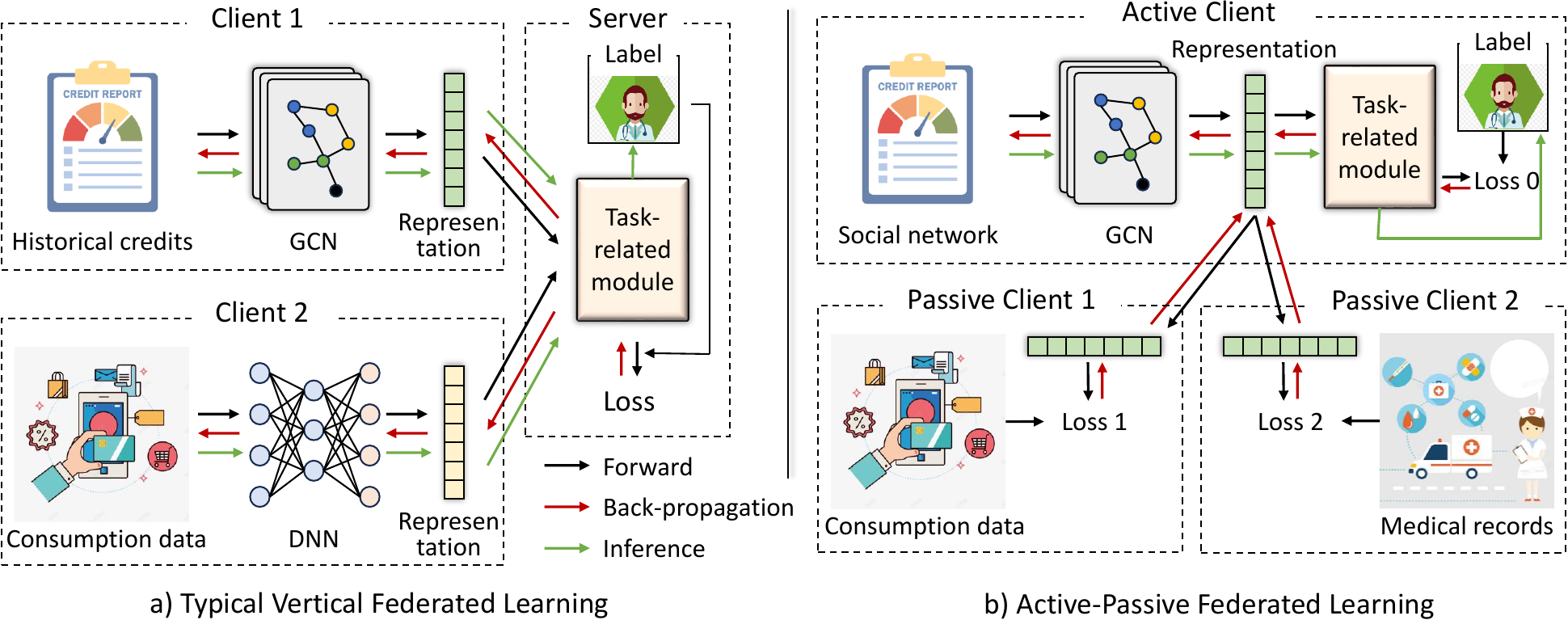}
    \caption{ Frameworks of typical vertical federated learning and the proposed APFed learning. For brevity, only two clients (the left), two passive clients (the right) and the classification task (the both sides) are presented, respectively. Meanwhile, Graph Convolutional Networks (GCN) and Deep Neural Networks (DNN) are employed to generated data representations. Actually, the client number, task type and representation generation method are not limited. Nevertheless, the arrows across clients and server refer to network transmission. Note that, the two data representations of the left are different (with different colors), while the three of the right are the same (with the same color).}
    \label{fig:vfl_apfl_framework}
\end{figure*}

% VFL scenario and framework
The mainstream methods of vertical federated learning \cite{DBLP:journals/tist/YangLCT19} follow the typical framework of Fig. \ref{fig:vfl_apfl_framework}.a. 
In the \emph{Forward} process of model training (black arrow), all clients firstly generate the data representations independently, then send them to the server so as to compute the loss with the task-related module centrally.
In the \emph{Back-propagation} process of model training (red arrow), the server computes the gradients with respect to each data representation, then separately transmits them to the clients, and each client can update the parameters of its representation generation method (GCN and DNN here). 
In the \emph{Inference} process of model training (green arrow), the procedure is the same with that of the \emph{Forward} process, plus an extra label prediction oriented from the task-related module.
To be summarized, vertical federated learning compulsorily requires the collaboration of all clients in both model training and inference, which we named \textbf{training together and inferring together} strategy.

% problem
However, the multi-view data and clients are usually possessed by different organizations.
In such case, it is promising for them to collaborate in model training, since the model training only needs a short period, i.e. hours or days.
On contrary, since the model inference is probably maintained for service in a long time, i.e. months or years, the collaboration is unpredictable due to a number of challenges, such as the concellation of contract, close down of organizations, network unavailability, offline deployment, etc., making the learned model invalid.

% APFed scenario and framework
To address this unpredictable collaboration problem, we identify a completely new and practical strategy, i.e. \textbf{training together but inferring solely} and propose a flexible Active-Passive Federated learning (APFed) framework, as shown in Fig. \ref{fig:vfl_apfl_framework}.b.
Specifically, the clients are separated to the active and the passive.
The active client is the initiator of a learning task, therefore, is unique and possesses a complete machine learning model, while the passive clients are its assistants.
In the \emph{Forward} process of model training, the active client generates the data representation and sends it to all passive clients.
Then, the passive clients regularize it with their data so as to compute corresponding losses.
In the \emph{Back-propagation} process, each passive client computes the gradients on the data representation and sends them back to the active client.
With unifying the gradients of both active and passive clients, the machine learning model can be updated.
Once the model is learned, the active client can make inference independently, avoiding the unpredictable collaboration problem between different organizations.
To validate the APFed framework, we also instance the APFed framework into two classification methods, i.e. APFed-R and APFed-C, with employing the reconstruction loss and the contrastive loss on passive clients, respectively.
Meanwhile, We conduct extensive experiments on the two methods and corresponding results well validate their effectiveness.

\section{Preliminary}
\label{sec:preliminary}

% vertical federated learning
Given $M$ clients $\{C_1, C_2,\cdots C_M\}$, the data of different features (multi-view data) are distributed across them with each identified by the unique sample IDs $\{I_1, I_2, \cdots, I_M\}$. 
In real-world applications, the sample IDs are not fully consistent but partially overlapped between clients, i.e. 
\begin{equation}\label{eq:sample_IDs_in_real-world}
I_1 \neq I_2 \neq \cdots \neq I_M, \quad I_1 \cap I_2 \cap \cdots \cap I_M = I \neq \emptyset.
\end{equation}
In a classical vertical federated learning framework, only the samples of set $I$ are concerned.
To achieve the sample ID alignment with preserving the data privacy, Private Set Intersection (PSI) protocols are mostly implemented and well studied in literature, such as \cite{DBLP:conf/eurocrypt/FreedmanNP04,DBLP:conf/uss/Pinkas0Z14,DBLP:conf/ndss/HuangEK12}.
In this paper, we apply PSI by default and only discuss the consequence after sample ID alignment. 

For the ease of expression, the aligned multi-view data and labels are denoted to be $\{\mathbf{x}_i^p\}_{i,p=1}^{N,M}$ and $\{\mathbf{y}_i\}_{i=1}^N$, respectively.
In existing researches, the data labels are always located on an independent server or one of the clients.
Note that, the federated learning model can be deployed on the both scenarios without modifying the loss objective and model architecture. 
Therefore, we organize the following contents in the former scenario (more common in literature) and denote the sever to be $S$.
In such setting, a vertical federated learning usually follows the three steps:

1) \emph{Forward}. The clients firstly encode their data into latent representations independently. 
Given the encoder parameters $\boldsymbol{\Theta}_p$ of client $C_p$, the latent representation $\mathbf{h}_i^p$ can be formulated as
\begin{equation}
\mathbf{h}_i^p = f_{\boldsymbol{\Theta}_p}(\mathbf{x}_i^p).
\end{equation}
Then, the obtained representations are transferred to the server. 
Denoting the task-related module and its parameters to be $g$ and $\boldsymbol{\Theta}_S$, the overall loss can be written as follows, 
\begin{equation}
L = \frac{1}{N}\sum_{i=1}^N\ell (g_{\boldsymbol{\Theta}_S}(\{\mathbf{h}_i^p\}_{p=1}^M), \mathbf{y}_i) + r_S(\boldsymbol{\Theta}_S) + \sum_{p=1}^Mr_p(\boldsymbol{\Theta}_p),
\end{equation}
in which $r_S(\cdot)$ and $r_p(\cdot)$ refer to the regularizations on the parameters of the server and $p$-th client.

\indent 2) \emph{Back-propagation}. The server firstly computes the gradients with regarding to each latent representation, i.e. $\partial\ell/\partial\mathbf{h}_i^p$, and transfers them back to each client.
Once the clients receive the gradients, their parameters can be updated. 

\indent 3) \emph{Inference}. For a test sample $\{\mathbf{x}_t^p\}_{p=1}^M$, each client is supposed to compute the its latent representation and transfers to the server so as to obtain the label
\begin{equation}
\mathbf{y}'_t = g_{\boldsymbol{\Theta}_S}(\{f_{\boldsymbol{\Theta}_p}(\mathbf{x}_t^p)\}_{p=1}^M).
\end{equation}
It is obvious that the label inference requires the collaborations of all clients.

\section{The proposed framework}
\label{sec:the_proposed_framework}

As seen in Section \ref{sec:preliminary}, the vertical federated learning requires the collaboration of all clients in not only the forward and back-propagation but also the inference steps, which summarizes to the \emph{training together and inferring together} strategy.
It is worth to note that, the multi-view data and clients are usually possessed by different organizations.
In such case, it is promising for them to collaborate in model training, due that the model training only needs a short period, i.e. hours or days.
However, since the model inference is probably maintained for service in a long time, i.e. months or years, the collaboration is unpredictable due to a number of challenges, such as the end of contract, close down of organizations, network unavailability, offline deployment, etc., making the learned model invalid.

To address the unpredictable collaboration problem in model inference, we propose a completely novel strategy, i.e. \emph{training together but inferring solely} and develop a flexible APFed framework which is visualized in Fig. \ref{fig:vfl_apfl_framework}.b.
Specifically, the active client is the initiator of a learning procedure and the only owner of the learned model, while the passive clients are its assistants.

With given the active client $C_A$ and the $p$-th passive client $C_p$ ($p\in\{1,2,\cdots,M\}$), we denote their data to be $\{\mathbf{x}_i^A\}_{i=1}^N$ and $\{\mathbf{x}_i^p\}_{i=1}^N$.
Note that the sample IDs obey Eq. (\ref{eq:sample_IDs_in_real-world}) and are aligned with PSI protocol by default.
In such case, the proposed APFed framework takes the following three steps:

1) \emph{Forward}. 
Firstly, the active client produces the latent representation of data sample $\mathbf{x}_i$ by
\begin{equation}\label{eq:latent_representation_apfed}
\mathbf{h}_i^A = f_{\boldsymbol{\Theta}_A}(\mathbf{x}_i^A),
\end{equation}
where $f_{\boldsymbol{\Theta}_A}(\cdot)$ is a parameterized mapping and is dependent on the data type.
Then, the active client $C_A$ distributes the generated representations $\{\mathbf{h}_i^A\}_{i=1}^N$ to all passive clients $\{C_1, C_2, \cdots, C_M\}$.
As a result, two parts of  loss are computed: 
a) In the active client, the latent representations are fed into the task-related module $g_{\boldsymbol{\Theta}_T}$, obtaining the loss as follows:
\begin{equation}\label{eq:loss_active_apfed}
L_A = \frac{1}{N}\sum_{i=1}^N \ell_A( g_{\boldsymbol{\Theta}_T}(\mathbf{h}_i^A), \mathbf{y}_i) + w_A\cdot r_A(\boldsymbol{\Theta}_A) + w_a\cdot r_a(\boldsymbol{\Theta}_T),
\end{equation}
in which $r_A(\cdot)$ and $r_a(\cdot)$ are the regularizations on neural network parameters. 
Correspondingly, $w_A$ and $w_a$ are their weight decay parameters.
b) The passive clients are supposed to compute the losses with their personal data and the received latent representations, i.e.
\begin{equation}\label{eq:loss_passive_apfed}
L_p = \frac{1}{N}\sum_{i=1}^N\ell_p(\mathbf{h}_i^A, \mathbf{x}_i^p; \boldsymbol{\Theta}_p) + w_p\cdot r_p(\boldsymbol{\Theta}_p).
\end{equation}
Similarly, $r_p$ and $w_p$ are the regularizations and weight decay parameter.
Finally, the overall loss can be presented as
\begin{equation}\label{eq:overall_loss_apfed}
L = L_A + \sum_{p=1}^M \lambda_p L_p,
\end{equation}
where $\lambda_p$ refers to the trade-off between the loss of active client and that of $p$-th passive client.
% Note that, only the latent representations $\{\mathbf{h}_i^A\}_{i=1}^N$ are shared among all clients, but the other information, such as model parameters, 

2) \emph{Back-propagation}.
To minimize the loss of Eq. (\ref{eq:overall_loss_apfed}), the gradient descent algorithm is employed.
At the very beginning, the gradients over the latent representation, i.e. $\mathbf{h}_i^A$ of Eq. (\ref{eq:latent_representation_apfed}), are computed on both active client $C_A$ and passive clients $\{C_1, C_2, \cdots, C_M\}$ separately.
Then, the passive clients send their gradients to the active one, obtaining 
\begin{equation}\label{eq:overall_gradients_apfed}
\frac{\partial L}{\partial \mathbf{h}_i^A} = \frac{\partial L_A}{\partial \mathbf{h}_i^A} + \sum_{p=1}^M \lambda_p \frac{\partial L_p}{\partial \mathbf{h}_i^A}.
\end{equation}
By back-propagating $\partial L / \partial \mathbf{h}_i^A$ on the active client, we can update the encoder parameters $\boldsymbol{\Theta}_A$.
Note that, apart from $\boldsymbol{\Theta}_A$, the parameters of the task-related module $\boldsymbol{\Theta}_T$ and those of the passive clients $\{\boldsymbol{\Theta}_p\}_{p=1}^M$ are also updated accordingly. 
Taking the widely used Stochastic Gradient Descent (SGD) strategy for an example, 
\begin{equation}\label{eq:Thetas_update_apfed}
    \boldsymbol{\Theta}_A = \boldsymbol{\Theta}_A - \alpha_A \frac{\partial L}{\partial \mathbf{h}_i^A}\cdot \frac{\partial \mathbf{h}_i^A}{\partial \boldsymbol{\Theta}_A}, \quad
    \boldsymbol{\Theta}_T = \boldsymbol{\Theta}_T - \alpha_T \frac{\partial L_A}{\partial \boldsymbol{\Theta}_T} \quad \text{and} \quad 
    \boldsymbol{\Theta}_p = \boldsymbol{\Theta}_p - \alpha_p \frac{\partial L_p}{\partial \boldsymbol{\Theta}_p},
\end{equation}
where $\alpha_A$, $\alpha_T$ and $\alpha_p$ denote the learning rates of corresponding modules and clients.

3) \emph{Inference}. 
After the \emph{Forward} and \emph{Back-propagation} steps, all the model parameters of the active client, including $\boldsymbol{\Theta}_A$ and $\boldsymbol{\Theta}_T$, are optimized and fixed.
For a test sample $\mathbf{x}_t^A$, the active client $C_A$ is able to produce the label as
\begin{equation}\label{eq:label_inference_apfed}
\mathbf{y}'_t = g_{\boldsymbol{\Theta}_A}(f_{\boldsymbol{\Theta}_A}(\mathbf{x}_t^A)).
\end{equation} 
It can be observed that the label $\mathbf{y}'_t$ can be computed on the active client independently, getting rid of the availability of data observations $\{\mathbf{x}_t^p\}_{p=1}^M$ and the collaboration of passive clients $\{C_1, C_2, \cdots, C_M\}$.

The overview of APFed framework is summarized in Alg. \ref{alg:pseudo_code_apfed}. 
To be summarized, the active client have the complete learning model to predict data labels in inference, while the passive clients only participate in model training, including forward and back-propagation.
Compared with the common vertical federated learning framework, the proposed APFed framework follows the \emph{training together but inferring solely} strategy and can deal with the unpredictable collaboration problem.

\begin{algorithm}[H]
\caption{Active-Passive Federated Learning Framework}
\label{alg:pseudo_code_apfed}
\vspace{5pt}
\begin{multicols}{2}
    \textbf{Training}\\
    * Active Client $C_A$
    \begin{algorithmic}[1]
    \STATE Align the data with the other clients via PSI;
    \STATE \emph{\# Forward}
    \STATE Compute the latent representations $\mathbf{h}_i^A$ via Eq. (\ref{eq:latent_representation_apfed});
    \STATE Send the latent representations $\mathbf{h}_i^A$ to all passive clients;
    \STATE Compute loss $L_A$ via Eq. (\ref{eq:loss_active_apfed});
    \STATE \emph{\# Back-propagation}
    \STATE Compute the gradients $\partial L_A / \partial \mathbf{h}_i^A$;
    \STATE Receive the gradients $\partial L_p / \partial \mathbf{h}_i^A$ from the passive clients;
    \STATE Compute the overall gradients $\partial L / \partial \mathbf{h}_i^A$ via Eq. (\ref{eq:overall_gradients_apfed});
    \STATE Update the parameters $\boldsymbol{\Theta}_T$ and $\boldsymbol{\Theta}_A$;
    \end{algorithmic}
    $\;$ \\
    * Passive Client $C_p$
    \begin{algorithmic}[1]
    \STATE Align data with the other clients via PSI;
    \STATE \emph{\# Forward}
    \STATE Receive latent representations $\mathbf{h}_i^A$ from the active client;
    \STATE Compute loss $L_p$ via Eq. (\ref{eq:loss_passive_apfed});
    \STATE \emph{\# Back-propagation}
    \STATE Compute the gradients $\partial L_p / \partial \mathbf{h}_i^A$;
    \STATE Send the gradients $\partial L_p / \partial \mathbf{h}_i^A$ to the active client;
    \STATE Update the parameters $\Theta_p$;
    \end{algorithmic}
    $\;$ \\
    $\;$ \\
    $\;$
\end{multicols} 
\vspace{-5pt}
\textbf{Inference}\\
* Active Client $C_A$;
\begin{algorithmic}[1]
\STATE Compute the label $\mathbf{y}'_t$ via Eq. (\ref{eq:label_inference_apfed}) independently without the collaboration of any passive clients;
\end{algorithmic}
\end{algorithm}

\section{The example methods}
\label{sec:the_example_methods}

In real-world scenarios, one can instance the proposed APFed framework of Section \ref{sec:the_proposed_framework} with different loss functions and neural network architectures. 
In this section, we implement two example methods by specifying those of active client and passive clients separately. 

As shown in Fig. \ref{fig:vfl_apfl_framework}, the active client can be obviously separated into two parts, i.e. the encoder $f_{\boldsymbol{\Theta}_A}$ and task-related module $g_{\boldsymbol{\Theta}_T}$.
In practice, one can simply instances them according to the data type and task, such as Convolutional Neural Networks (CNN) and MultiLayer Perceptron (MLP) for image classification.

As for the passive clients, we give out two practical example loss functions:

1) \emph{Reconstruction loss}.
After receiving the latent representation $\mathbf{h}_i^A$ of the $i$-th data sample, the $p$-th passive client can use it to reconstruct data in $p$-th view.
In this way, the loss $\ell_p$ of Eq. (\ref{eq:loss_passive_apfed}) is formulated as 
\begin{equation}\label{eq:reconstruction_loss}
\ell_p(\mathbf{h}_i^A, \mathbf{x}_i^p; \boldsymbol{\Theta}_p) = L_q(\mathbf{x}_i^p - DEC_{\boldsymbol{\Theta}_p}(\mathbf{h}_i^A)).
\end{equation}
Here, $DEC_{\boldsymbol{\Theta}_p}$ is the parameterized decoder and $L_q$ indicates the vector norm, such as the widely-used $L_2$-norm.
Note that, they are mostly customized in accordance with data type and preference of $p$-th passive client.

2) \emph{Contrastive loss}.
Different from the reconstruction loss, the contrastive loss is required to compute on data batches. 
Given an arbitrary data batch $\{\mathbf{x}_i^p\}_{i,p=1}^{N',M}$, the $p$-th passive client firstly encodes its data samples $\{\mathbf{x}_i^p\}_{i}^{N'}$ into latent representations, i.e.
\begin{equation}\label{eq:encoder_contrastive_loss}
\mathbf{h}_i^p = ENC_{\boldsymbol{\Theta}_p}(\mathbf{x}_i^p), 
\end{equation}
in which $ENC_{\boldsymbol{\Theta}_p}$ is the parameterized encoder and, as well, customized in accordance with the preference of $p$-th passive client.
With the latent representations $\{\mathbf{h}_i^p\}_{i}^{N'}$ and $\{\mathbf{h}_i^A\}_{i}^{N'}$, contrastive learning \cite{DBLP:journals/ijcv/LiYPLHP22} considers those of the data sample from different clients to be positive pairs while the others to be negative pairs.
By maximizing the similarity of positive pairs and minimizing that of negative pairs, the contrastive loss of $i$-th data sample can be calculated by following 
\begin{equation}
\begin{split}
\hat{\ell}_{i,p} =  - \log \frac{\exp(s(\mathbf{h}_i^A, \mathbf{h}_i^p)/\lambda)}{\sum_{j=1}^{N'} [ \mathbbm{1}_{j\neq i} \exp(s(\mathbf{h}_i^A, \mathbf{h}_j^A)/\lambda) + \exp(s(\mathbf{h}_i^A, \mathbf{h}_j^p)/\lambda)]}, 
\end{split}
\end{equation}
where $\mathbbm{1}_{j\neq i}\in\{0, 1\}$ is the indicator function and set to $0$ if and only if $j=i$.
At the same time, $\tau$ denotes a temperature parameter and $s(\cdot, \cdot)$ measures the similarity between vectors with
\begin{equation}
s(\mathbf{h}, \mathbf{h}') = \frac{\mathbf{h}^\top\mathbf{h}'}{\|\mathbf{h}\|\|\mathbf{h}'\|}.
\end{equation}
Finally, the overall contrastive loss is implemented as 
\begin{equation}\label{eq:contrastive_loss}
  \ell_p(\mathbf{h}_i^A, \mathbf{x}_i^p; \boldsymbol{\Theta}_p) = \frac{1}{N'} \sum_{i=1}^{N'} \hat{\ell}_{i,p}.
\end{equation}

Based on the above implementation instructions, we develop two novel classification methods named APFed-R and APFed-C by employing the reconstruction loss and the contrastive loss in passive clients, respectively.
Their specifications are summarized in Table \ref{tab:specifications_example_methods}.

\begin{table}[t]
    \caption{The specifications of the APFed-R and APFed-C methods.}
    \label{tab:specifications_example_methods}
    \centering
    \renewcommand\arraystretch{1.3}
    \begin{tabular}{llcc}
    \toprule 
    Client & Item & APFed-R & APFed-C \\ \midrule 
    \multirow{2}{*}{Active - Eq. (\ref{eq:loss_active_apfed})} & $\ell_A$ & Cross Entropy Loss &  Cross Entropy Loss\\ 
     & $r_A$, $r_a$ & $L_2$-norm & $L_2$-norm \\ \midrule 
    \multirow{2}{*}{Passive - Eq. (\ref{eq:loss_passive_apfed})} & $\ell_p$ & Eq. (\ref{eq:reconstruction_loss}) & Eq. (\ref{eq:contrastive_loss}) \\
     & $r_p$ & $L_2$-norm & $L_2$-norm \\ 
    \bottomrule
    \end{tabular}
\end{table}

\section{Experiment}
\label{sec:experiment}

In order to validate the effectiveness of the proposed APFed framework, we conduct extensive experiments on the two example methods in Section \ref{sec:the_example_methods}.

\subsection{Experiment setting}
\label{subsec:experiment_setting}

\begin{figure}[t]
  \centering
  \includegraphics[width=\linewidth]{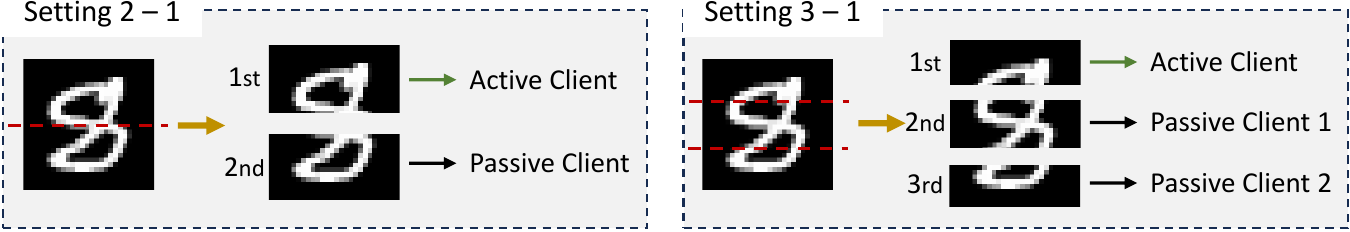}
  \caption{The $m$-$i$ splitting strategy on MNIST dataset. In the 2-1 setting, an "8" image is cut into 3 parts (views) with deploying the 1-st view into the active client and the 2-nd view to the passive client. Similar is the 3-1 setting. \\ \vspace{-10pt}}
  \label{fig:data_split}
\end{figure}

In the experiments, we adopt four popular image benchmarks to validate the proposed framework and methods, including MNIST, Fashion MNIST (FMNIST), CIFAR10 and CIFAR100. 
To make the single-view data into multi-view setting, we follow the splitting strategy of most vertical federated learning researches \cite{DBLP:journals/corr/abs-2007-06081}, in which the images are cut into $m$ parts to be $m$ data views.
For simplicity, we split all the images horizontally in the following experiments. 
Note that splitting them vertically also works without any modifications apart from the architectures of encoding and decoding neural networks.
On the basis, we develop the APFed settings with each being recorded as $m$-$i$, indicating the $i$-view data is deployed on the active client while the rest views are on passive clients. 
Besides, the labels are all deployed on the active client. 
To be more comprehensive, we show the example settings on MNIST dataset in Fig. \ref{fig:data_split}.

\begin{table*}[!t]
    \centering
    \small
    \caption{Accuracy comparison of the proposed APFed-R, APFed-C and comparative baselines. Note that, the best results are marked in bold, while the second best underlined.}
    \label{tab:accuracy_comparison_with_baselines}
    \renewcommand\arraystretch{1.1}
    \setlength\tabcolsep{5.5pt}
    \begin{tabular}{llcccccc}
    \toprule
    Dataset & Setting & TVFL-0 & TVFL-A & TVFL-R & SingleVL & APFed-R & APFed-C \\ \midrule
    \multirow{6}{*}{MNIST} & 2 - 1 & 52.41{\scriptsize $\pm$ 2.77} & 72.69{\scriptsize $\pm$ 1.28} & 10.42{\scriptsize $\pm$ 0.14} & 95.24{\scriptsize $\pm$ 0.12} & \underline{95.52}{\scriptsize $\pm$   0.07} & \textbf{95.72}{\scriptsize $\pm$ 0.09} \\
     & 2 - 2 & 49.01{\scriptsize $\pm$ 1.64} & 55.29{\scriptsize $\pm$ 2.94} & 11.32{\scriptsize $\pm$ 1.02} & 92.12{\scriptsize $\pm$ 0.16} & \underline{92.38}{\scriptsize $\pm$   0.10} & \textbf{92.93}{\scriptsize $\pm$ 1.33} \\
     & 3 - 1 & 22.22{\scriptsize $\pm$ 1.30} & 21.33{\scriptsize $\pm$ 1.72} & 8.37{\scriptsize $\pm$ 2.01} & 78.80{\scriptsize $\pm$ 0.15} & \textbf{79.77}{\scriptsize $\pm$   0.14} & \underline{79.13}{\scriptsize $\pm$ 0.22} \\
     & 3 - 2 & 35.13{\scriptsize $\pm$ 1.59} & 63.99{\scriptsize $\pm$ 2.57} & 11.89{\scriptsize $\pm$ 3.61} & 95.58{\scriptsize $\pm$ 0.06} & \underline{96.12}{\scriptsize $\pm$   0.09} & \textbf{96.29}{\scriptsize $\pm$ 0.96} \\
     & 3 - 3 & 21.86{\scriptsize $\pm$ 1.30} & 29.18{\scriptsize $\pm$ 2.86} & 10.53{\scriptsize $\pm$ 1.19} & 79.41{\scriptsize $\pm$ 0.19} & \underline{80.22}{\scriptsize $\pm$   0.08} & \textbf{80.30}{\scriptsize $\pm$ 1.98} \\ 
     & Avg. & 36.13 & 48.50 & 10.51 & 88.23 & \underline{88.80} & \textbf{88.87} \\ \midrule
    \multirow{6}{*}{FMNIST} & 2 - 1 & 68.22{\scriptsize $\pm$ 1.72} & 70.61{\scriptsize $\pm$ 1.49} & 11.61{\scriptsize $\pm$ 2.64} & 88.49{\scriptsize $\pm$ 0.16} & \textbf{88.94}{\scriptsize $\pm$   0.25} & \underline{88.85}{\scriptsize $\pm$ 0.63} \\
     & 2 - 2 & 60.08{\scriptsize $\pm$ 1.52} & 55.41{\scriptsize $\pm$ 2.09} & 10.49{\scriptsize $\pm$ 0.68} & 87.62{\scriptsize $\pm$ 0.16} & \textbf{88.31}{\scriptsize $\pm$   0.14} & \underline{88.13}{\scriptsize $\pm$ 1.39} \\
     & 3 - 1 & 45.58{\scriptsize $\pm$ 1.10} & 41.19{\scriptsize $\pm$ 1.21} & 10.00{\scriptsize $\pm$ 0.00} & 80.54{\scriptsize $\pm$ 0.23} & \underline{81.38}{\scriptsize $\pm$   0.22} & \textbf{81.57}{\scriptsize $\pm$ 0.24} \\
     & 3 - 2 & 52.15{\scriptsize $\pm$ 3.57} & 46.12{\scriptsize $\pm$ 1.85} & 11.98{\scriptsize $\pm$ 1.36} & 86.19{\scriptsize $\pm$ 0.16} & \textbf{87.09}{\scriptsize $\pm$   0.18} & \underline{86.65}{\scriptsize $\pm$ 0.21} \\
     & 3 - 3 & 43.83{\scriptsize $\pm$ 3.99} & 32.75{\scriptsize $\pm$ 1.65} & 10.10{\scriptsize $\pm$ 0.20} & 84.50{\scriptsize $\pm$ 0.20} & \textbf{85.43}{\scriptsize $\pm$   0.15} & \underline{85.40}{\scriptsize $\pm$ 0.34} \\ 
     & Avg. & 53.97 & 49.22 & 10.84 & 85.47 & \textbf{86.23} & \underline{86.12} \\ \midrule
    \multirow{6}{*}{CIFAR10} & 2 - 1 & 43.24{\scriptsize $\pm$ 0.85} & 40.11{\scriptsize $\pm$ 1.00} & 12.09{\scriptsize $\pm$ 0.57} & 52.94{\scriptsize $\pm$ 1.36} & \underline{53.74}{\scriptsize $\pm$ 1.35} & \textbf{55.99}{\scriptsize $\pm$ 2.85} \\
     & 2 - 2 & 48.91{\scriptsize $\pm$ 1.52} & 44.99{\scriptsize $\pm$ 0.81} & 13.52{\scriptsize $\pm$ 1.29} & 56.53{\scriptsize $\pm$ 2.20} & \underline{57.05}{\scriptsize $\pm$   1.44} & \textbf{59.50}{\scriptsize $\pm$ 0.48} \\
     & 3 - 1 & 21.61{\scriptsize $\pm$ 2.43} & 22.04{\scriptsize $\pm$ 1.71} & 10.79{\scriptsize $\pm$ 0.29} & \underline{45.07}{\scriptsize $\pm$ 0.79} & 44.66{\scriptsize $\pm$   0.63} & \textbf{46.93}{\scriptsize $\pm$ 0.51} \\
     & 3 - 2 & 39.07{\scriptsize $\pm$ 1.28} & 33.06{\scriptsize $\pm$ 1.36} & 12.11{\scriptsize $\pm$ 1.05} & 54.17{\scriptsize $\pm$ 1.29} & \underline{55.03}{\scriptsize $\pm$   0.51} & \textbf{56.02}{\scriptsize $\pm$ 0.80} \\
     & 3 - 3 & 40.93{\scriptsize $\pm$ 1.72} & 33.36{\scriptsize $\pm$ 1.39} & 11.44{\scriptsize $\pm$ 0.70} & 52.60{\scriptsize $\pm$ 1.65} & \underline{54.11}{\scriptsize $\pm$   0.29} & \textbf{55.56}{\scriptsize $\pm$ 0.65} \\ 
     & Avg. & 38.75 & 34.71 & 11.99 & 52.26 & \underline{52.92} & \textbf{54.80} \\ \midrule
    \multirow{6}{*}{CIFAR100} & 2 - 1 & 11.94{\scriptsize $\pm$ 0.09} & 10.83{\scriptsize $\pm$ 0.56} & 1.30{\scriptsize $\pm$ 0.11} & 21.04{\scriptsize $\pm$ 0.27} & \underline{21.65}{\scriptsize $\pm$   0.50} & \textbf{24.31}{\scriptsize $\pm$ 0.36} \\
     & 2 - 2 & 10.51{\scriptsize $\pm$ 0.84} & 9.06{\scriptsize $\pm$ 1.66} & 1.26{\scriptsize $\pm$ 0.14} & 20.38{\scriptsize $\pm$ 0.34} & \underline{21.42}{\scriptsize $\pm$   0.65} & \textbf{23.72}{\scriptsize $\pm$ 0.63} \\
     & 3 - 1 & 5.76{\scriptsize $\pm$ 0.55} & 6.05{\scriptsize $\pm$ 0.72} & 1.13{\scriptsize $\pm$ 0.10} & 18.27{\scriptsize $\pm$ 0.46} & \underline{19.38}{\scriptsize $\pm$   0.26} & \textbf{20.09}{\scriptsize $\pm$ 0.37} \\
     & 3 - 2 & 11.81{\scriptsize $\pm$ 1.00} & 9.48{\scriptsize $\pm$ 0.84} & 1.33{\scriptsize $\pm$ 0.09} & 22.80{\scriptsize $\pm$ 0.46} & \underline{23.84}{\scriptsize $\pm$   0.34} & \textbf{25.99}{\scriptsize $\pm$ 1.09} \\
     & 3 - 3 & 6.18{\scriptsize $\pm$ 0.52} & 4.60{\scriptsize $\pm$ 0.34} & 1.25{\scriptsize $\pm$ 0.26} & 19.78{\scriptsize $\pm$ 0.33} & \underline{21.29}{\scriptsize $\pm$   0.42} & \textbf{22.37}{\scriptsize $\pm$ 0.66} \\ 
     & Avg. & 9.24 & 8.00 & 1.25 & 20.45 & \underline{21.52} & \textbf{23.30} \\ \bottomrule
    \end{tabular}
\end{table*}

Since the proposed APFed framework is the first attempt to address the unpredictable collaboration problem of federated learning, we consider the typical vertical federated learning framework \cite{DBLP:journals/corr/abs-2007-06081,10415268,DBLP:journals/corr/abs-2304-01829} (abbr. TVFL whose framework is shown in Fig. \ref{fig:vfl_apfl_framework}.a) as baseline and directly compare them empirically.
It is obvious that they differ each other from the model inference, where the former only adopts the active client, while the latter requires all the clients.
To simulate the unpredictable collaboration consequence in inference and ensure the fairness of comparison, we use only one client of TVFL in inference by compulsorily setting the latent representations of the other clients to zero, average value and random value, obtaining another three methods: TVFL-0, TVFL-A and TVFL-R, respectively.
Nevertheless, to provide a strong baseline, we also develop the Single-view Learning (SingleVL) method in which the model training and inference are both completed on one client solely. 
In other words, it is only the active client of the proposed APFed framework without any passive clients.

To ensure the fairness, we use the same backbone neural networks in the aforementioned methods, as well, unify their parameters, such as learning rate, weight decay, etc.
Nevertheless, the codes of the proposed APFed-R, APFed-C and competing baselines are available at Github (open once accepted).

\subsection{Comparison to baselines}
\label{subsec:comparison-to-baselines}

To validate the effectiveness of the proposed APFed framework, we compare two of its implementations, i.e. APFed-R and APFed-C, with four baselines, including TVFL-0, TVFL-A, TVFL-R and SingleVL.
Corresponding experiment results are shown in Table \ref{tab:accuracy_comparison_with_baselines}.

The former three TVFL baselines simulate the consequence where the unpredictable collaboration problem is not considered in model training. 
Concretely, they follow the training procedure of Fig. \ref{fig:vfl_apfl_framework}.a.
Due that, in APFed setting, the active client should infer the label of incoming data independently, they replace the representations from passive clients with self-generated ones, i.e. zeros, average of the active client's representations and random values. 
It can be observed that the three methods achieve bad results in all settings and datasets. 
Especially, TVFL-R is the worst and close to random guess (10\%, 10\%, 10\% and 1\% on four datasets, respectively). 
This indicates that the typical vertical federated learning methods are vulnerable to the unpredictable collaboration problem, hence illustrating the motivation rationality of this research. 
On contrary, the proposed APFed-R and APFed-C increase the accuracies by multiple times according to all settings, well validating the effectiveness of them and the APFed framework. 

Apart from the TVFL baselines, SingleVL simulates the consequence where the active client trains a single-view model as back-up and use it once the collaboration broken. 
As seen from the table, both APFed-R and APFed-C outperforms SingleVL in almost all settings, except one slight decrease, i.e. 0.41\% (3-1 setting on CIFAR10).
On average, the former APFed-R outperforms the latter by 0.57\%, 0.76\%, 0.66\% and 1.07\%, while the former APFed-C by 0.64\%, 0.65\%, 2.54\% and 2.85\%, respectively.
As for the comparison between APFed-R and APFed-C, we will make a discussion in Section \ref{sec:discussion}.

Overall, it can be concluded that the proposed APFed-R and APFed-C outperform the baselines empirically, well validating the effectiveness of the proposed APFed framework. 

\begin{figure*}[t]
    \centering
    \includegraphics[width=\linewidth]{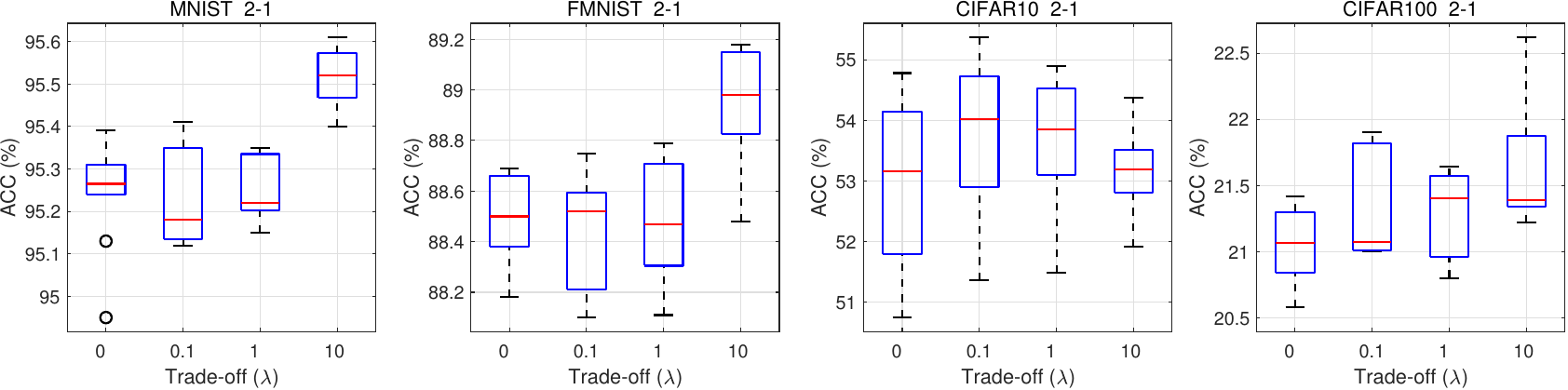} \\ \vspace{-1pt}
    \caption{The accuracy comparison of the proposed \textbf{APFed-R} method with trade-off $\lambda$ ranging from 0 to 10 in the 2-1 setting.}
    \label{fig:trade_off_dec}
\end{figure*}
\begin{figure*}[!t]
    \centering
    \includegraphics[width=\linewidth]{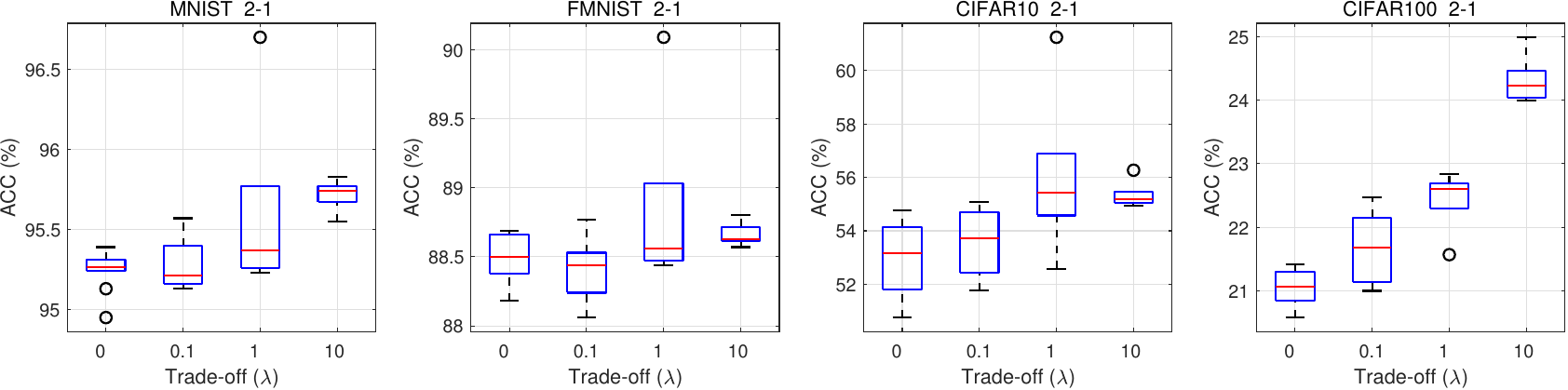} \\ \vspace{-1pt}
    \caption{The accuracy comparison of the proposed \textbf{APFed-C} method with trade-off $\lambda$ ranging from 0 to 10 in the 2-1 setting.}
    \label{fig:trade_off_con}
\end{figure*}

\subsection{Parameter study}
\label{subsec:parameter-study}

Apart from the parameters of neural networks, the proposed APFed framework requires to incorporate the gradients from passive clients, hence introduces an extra hyper-parameter $\lambda$.
Therefore, we investigate the results of APFed-R and APFed-C methods with $\lambda$ ranging from 0 to 10.
Due to the space limit, the accuracies on MNIST and CIFAR10 are shown in Fig. \ref{fig:trade_off_dec} and \ref{fig:trade_off_con}. 
Two observations are listed in the following.
\begin{enumerate}[1)]
  \item It is obvious that both of the APFed methods degrade to SingleVL when parameter $\lambda$ setting to 0. 
  It can be found that the accuracies when $\lambda \neq 0$ are better than those when $\lambda = 0$ in all settings.
  This demonstrates incorporating the active client benefits from the information of passive clients, validating the effectiveness of the proposed APFed framework again.
  \item Looking into the consequences with $\lambda$ ranging from 0.1 to 10, we observe that the accuracies of both APFed-R and APFed-C increase consistently on MNIST, FMNIST and CIFAR100 when $\lambda$ taking a larger value. 
  At the same time, one different observation is obtained on CIFAR10, where the accuracies decrease slightly when $\lambda$ increasing from 1 to 10.
  To be summarized, small parameter $\lambda$ results in a relatively low performance, and it is supposed to be set to 1 or a larger value in further practical use.
  Nevertheless, the phenomenon empirically proves both the reconstruction loss and the contrastive loss of passive clients are effective to help the active client learn a better model. 
\end{enumerate}

% \subsection{Insights of the model training}
% \label{subsec:insight-model-training}

% In the following, we provide a detailed insight of model training by showing the accuracy variation and loss of the active client along with the optimization process in Fig. \ref{fig:insight_apfed_methods}.
% It can be seen that the accuracies of the APFed-R and APFed-C methods increase consistently at the beginning, then converge at their maximums between 100 to 200 epochs. 
% Meanwhile, their losses decrease dramatically at first and then converge to the minimums.
% The two aforementioned observations is in accord with the training phenomenon of most neural networks, such as that of SingleVL. 
% By comparing the two proposed methods with the SingleVL baseline, the former achieve obvious improvements over the latter on accuracy, which is consist with the results in Table \ref{tab:accuracy_comparison_with_baselines}.
% Besides, one interesting observation is investigated that APFed-C achieves a faster dropping rate and a smaller value of the active client's loss than SingleVL. 
% To be more specific, SingleVL is only required to minimize the active client's loss, while APFed-C is required to minimize both of that and the passive clients' losses, i.e. contrastive loss in Eq. (\ref{eq:contrastive_loss}). 
% This, in combination with the accuracy improvement of APFed-C, well illustrates that the participation of passive clients could help the active client minimize the its loss, thereby optimize its neural network parameters closer to the global optimum.

\section{Discussion}
\label{sec:discussion}

In the former section, we validate the effectiveness of APFed framework via conducting a set of experiments on the APFed-R and APFed-C methods. 
Here, we continue to make the following four discussions.

1) \emph{APFed-R v.s. APFed-C}.
By comparing the accuracies of the two example methods in Table \ref{tab:accuracy_comparison_with_baselines}, APFed-C achieve better results on three datasets, especially on CIFAR10 and CIFAR100, while APFed-R is better only on FMNIST. 
Also, APFed-R may be partially limited by the design and implementation difficulties of reconstruction neural networks, i.e. $DEC_{\boldsymbol{\Theta}_p}(\cdot)$ in Eq. (\ref{eq:reconstruction_loss}).
For example, reconstructing a segment of audio from its latent representations requires a large amount labor effort. 
On contrary, APFed-C is dependent on encoders, as seen from Eq. (\ref{eq:encoder_contrastive_loss}), and there are plenty of well designed or even pretrained encoders in communities, significantly reducing its implementation difficulty. 
Apart from the aforementioned advantages, compared with APFed-R, APFed-C is weaker on training efficiency especially when large batch size is adopted (the computation complexity of its contrastive loss is square of data batch size).
Therefore, one should choose APFed-R and APFed-C according to the target application setting.

2) \emph{Blackbox of the passive client}.
In the setting of APFed framework, active client is the initiator of a learning task and possesses the complete learned model.
It only communicates with the passive clients by sending the encoded latent representations and receiving the corresponding gradients, but leaves the way to produce gradients from latent representations without any limitations.
Hence, the passive clients can be or are always blackboxes respect to the active client, undoubtedly enhancing their willingness of participation.

3) \emph{Flexibility}.
The APFed framework is highly flexible. 
Due to the blackbox property mentioned above, the passive clients can independently customize the neural networks according to their data formats and  requirements.
Nevertheless, their loss functions are not only limited to the reconstruction loss and the contrastive loss defined in Section \ref{sec:the_example_methods}.
Some other functions, such as multi-view clustering loss \cite{DBLP:journals/tsmc/WenZFZXZL23,DBLP:conf/cvpr/Wen0X0HF023}, are also feasible.
Besides, different passive clients can employ different loss functions to assistant the active client for a same task.

4) \emph{Limitation}.
This paper may be partially limited from the following two aspects. 
First, all experiments are conducted on the image classification task, leaving the benefits of the proposed APFed framework on other tasks not validated. 
Moreover, contributions of the passive clients are not quantified, which may be a potential obstacle of collaboration.
In the future work, we will try to implement the APFed framework on more downstream tasks, such as multi-model object detection, tracking, anomaly detection, etc., and develop a practical mechanism to quantify the contributions of passive clients.

\section{Conclusion}
\label{sec:conclusion}

Existing vertical federated learning methods requires the collaboration of all clients in both machine learning model training and inference, resulting in the failures when partial clients are not available. 
To address the problem, this paper proposes a flexible Active-Passive Federated learning (APFed) framework, to the first attempt, making model inference with only one client independently. 
Besides, two example methods, i.e. APFed-R and APFed-C, are implemented and validated in experiments. 
Corresponding results well proves the effectiveness of the proposed framework and methods, along with exploring some of their properties.

{\small
\bibliographystyle{unsrt}
\bibliography{ref.bib}
}

%%%%%%%%%%%%%%%%%%%%%%%%%%%%%%%%%%%%%%%%%%%%%%%%%%%%%%%%%%%%

\appendix

\section{Appendix}

\subsection{Experiment details}
\label{subsec:exp_details_appendix}

\emph{Dataset}. 
This paper employs four public and widely-used datasets in experiments. 
They are 
MNIST (\url{http://yann.lecun.com/exdb/mnist/}), 
FMNIST (Fashion MNIST, \url{https://github.com/zalandoresearch/fashion-mnist}), 
CIFAR10 (\url{https://www.cs.toronto.edu/~kriz/cifar.html}) and 
CIFAR100 (\url{https://www.cs.toronto.edu/~kriz/cifar.html}), 
where FMNIST is licenced by The MIT License (MIT) Copyright © [2017] Zalando SE (\url{https://tech.zalando.com}), while the rest are publicly used without licences.

\emph{Harware}.
Our experiments are performed on a server with two 2.40GHz 20-core Intel Xeon Silver 4210R CPUs and two Nvidia 3090 GPUs.

\emph{Neural network settings}.
For ease of reproducibility, the neural network settings of the proposed APFed-R and APFed-C methods are listed in Table \ref{tab:net_settings_mnist} and \ref{tab:net_settings_cifar}.
Note that, the two tables are organized by following the rules:
1) conv/deconv$i$ ($in\_channel$, $out\_channel$, $kernel\_size$): $i$-th parameterized convolutional/deconvolutional layer;
2) fc$i$ ($in\_dimension$, $out\_dimension$): $i$-th parameterized fully-connected layer;
3) $d$ is dependent on the size of images.

\begin{table}[h]
    \caption{The neural network settings of the APFed-R and APFed-C methods on MNIST and FMNIST.}
    \label{tab:net_settings_mnist}
    \centering
    \renewcommand\arraystretch{1.3}
    \begin{tabular}{llcc}
    \toprule 
    Client & Network & APFed-R & APFed-C \\ \midrule 
    \multirow{2}{*}{Active} & $f_{\boldsymbol{\Theta}_A}$ & \makecell{conv1 ($\;\;$1, 31, 5) \\ conv2 (32, 64, 5)} &  \makecell{conv1 ($\;\;$1, 31, 5) \\ conv2 (32, 64, 5)}\\  \cline{2-4} 
     & $g_{\boldsymbol{\Theta}_T}$ & \makecell{fc1 ($\;\;$$d$, 256) \\ fc2 (256, 10)} & \makecell{fc1 ($\;\;$$d$, 256) \\ fc2 (256, 10)} \\ \midrule 
    \multirow{2}{*}{Passive} & $DEC_{\boldsymbol{\Theta}_p}$ & \makecell{deconv1 (64, 32, 5) \\ deconv2 (32, $\;\;$1, 5)} & - \\ \cline{2-4}
     & $ENC_{\boldsymbol{\Theta}_p}$ & - & \makecell{conv1 ($\;\;$1, 31, 5) \\ conv2 (32, 64, 5)} \\ \midrule
     \multirow{3}{*}{All} & optimizer & SGD (momentum 0.9) & SGD (momentum 0.9) \\ 
     & learning rate & $1e^{-3}$ & $1e^{-3}$ \\ 
     & weight decay & $1e^{-4}$ & $1e^{-4}$ \\ \bottomrule
    \end{tabular}
\end{table}

\begin{table}[h]
    \caption{The neural network settings of the APFed-R and APFed-C methods on CIFAR10 and CIFAR100.}
    \label{tab:net_settings_cifar}
    \centering
    \renewcommand\arraystretch{1.3}
    \begin{tabular}{llcc}
    \toprule 
    Client & Network & APFed-R & APFed-C \\ \midrule 
    \multirow{2}{*}{Active} & $f_{\boldsymbol{\Theta}_A}$ & \makecell{conv1 ($\;\;$3, 16, 5) \\ conv2 (16, 32, 4) \\ conv3 (32, 64, 5)} &  \makecell{conv1 ($\;\;$3, 16, 5) \\ conv2 (16, 32, 4) \\ conv3 (32, 64, 5)}\\  \cline{2-4} 
     & $g_{\boldsymbol{\Theta}_T}$ & \makecell{fc1 ($\;\;$$d$, 50) \\ fc2 (50, 10/100)} & \makecell{fc1 ($\;\;$$d$, 50) \\ fc2 (50, 10/100)} \\ \midrule 
    \multirow{2}{*}{Passive} & $DEC_{\boldsymbol{\Theta}_p}$ & \makecell{deconv1 (64, 32, 5) \\ deconv2 (32, 16, 4) \\ deconv3 (16, $\;\;$3, 5)} & - \\ \cline{2-4}
     & $ENC_{\boldsymbol{\Theta}_p}$ & - & \makecell{conv1 ($\;\;$3, 16, 5) \\ conv2 (16, 32, 4) \\ conv3 (32, 64, 5)} \\ \midrule
     \multirow{3}{*}{All} & optimizer & SGD (momentum 0.9) & SGD (momentum 0.9) \\ 
     & learning rate & $1e^{-3}$ & $1e^{-3}$ \\ 
     & weight decay & $1e^{-4}$ & $1e^{-4}$ \\ \bottomrule
    \end{tabular}
\end{table}

\end{document}